\documentclass[runningheads]{llncs}

\usepackage[year=2026]{eccv}

\usepackage{eccvabbrv}
\usepackage{wrapfig}
\usepackage{graphicx}
\usepackage{booktabs}
\usepackage{multirow}
\usepackage{makecell}
\usepackage{colortbl}

\usepackage[accsupp]{axessibility}  
\usepackage{wrapfig} 

\usepackage{hyperref}

\usepackage{orcidlink}

\begin{document}

\title{Decoupling Perception from Reasoning for Hallucination-Resistant Video Understanding} 

\titlerunning{Abbreviated paper title}

\author{
Bowei Pu\inst{1} \and
Chuanbin Liu\inst{1}\thanks{Corresponding author.} \and
Yifan Ge\inst{1} \and
Peicheng Zhou\inst{1} \and
Yiwei Sun\inst{1} \and
Zhiying Lu\inst{1} \and
Zhangchi Hu\inst{1} \and
Hongtao Xie\inst{1}
}

\authorrunning{B. Pu et al.}

\institute{
University of Science and Technology of China, Hefei, China \\
\email{\{pubowei\}@mail.ustc.edu.cn, \{liucb92, htxie\}@ustc.edu.cn}
}

\maketitle
\begin{abstract}
Video Large Language Models improve reasoning over complex videos by generating intermediate reasoning text.  However, reliable reasoning depends on accurate video perception. In existing approaches, perception evidence is intertwined with reasoning text, making it difficult to directly supervise the perception process.
We argue that reliable supervision requires explicitly separating perception evidence from reasoning so that perception can be verified independently.
To supervise perception directly, we propose \textbf{Decoupled Perception and Logic (DPL)}, which represents perception as fixed-format evidence units containing timestamps and visual descriptions. This structured representation enables direct extraction of perception content and simplifies alignment between video segments and reward evaluation.
Building on DPL, we introduce a perception reward that encourages both hallucination resistance and perception-based reasoning. An \textbf{Factual-Aware Evaluator (FAE)} provides anti-hallucination scores and achieves hallucination evaluation performance comparable to GPT-4o. In addition, we validate reasoning \textbf{consistency} by feeding perception results and questions into a reference model.
Experiments show that, by providing reliable process rewards, Video-DPL consistently improves post-training performance at both 3B and 7B scales, while delivering higher data efficiency.
\keywords{Video Reasoning \and Video Hallucination}
\end{abstract}

    \section{Introduction}
    \label{sec:intro}
    \begin{figure}[t]
        \centering
        \includegraphics[width=1\linewidth]{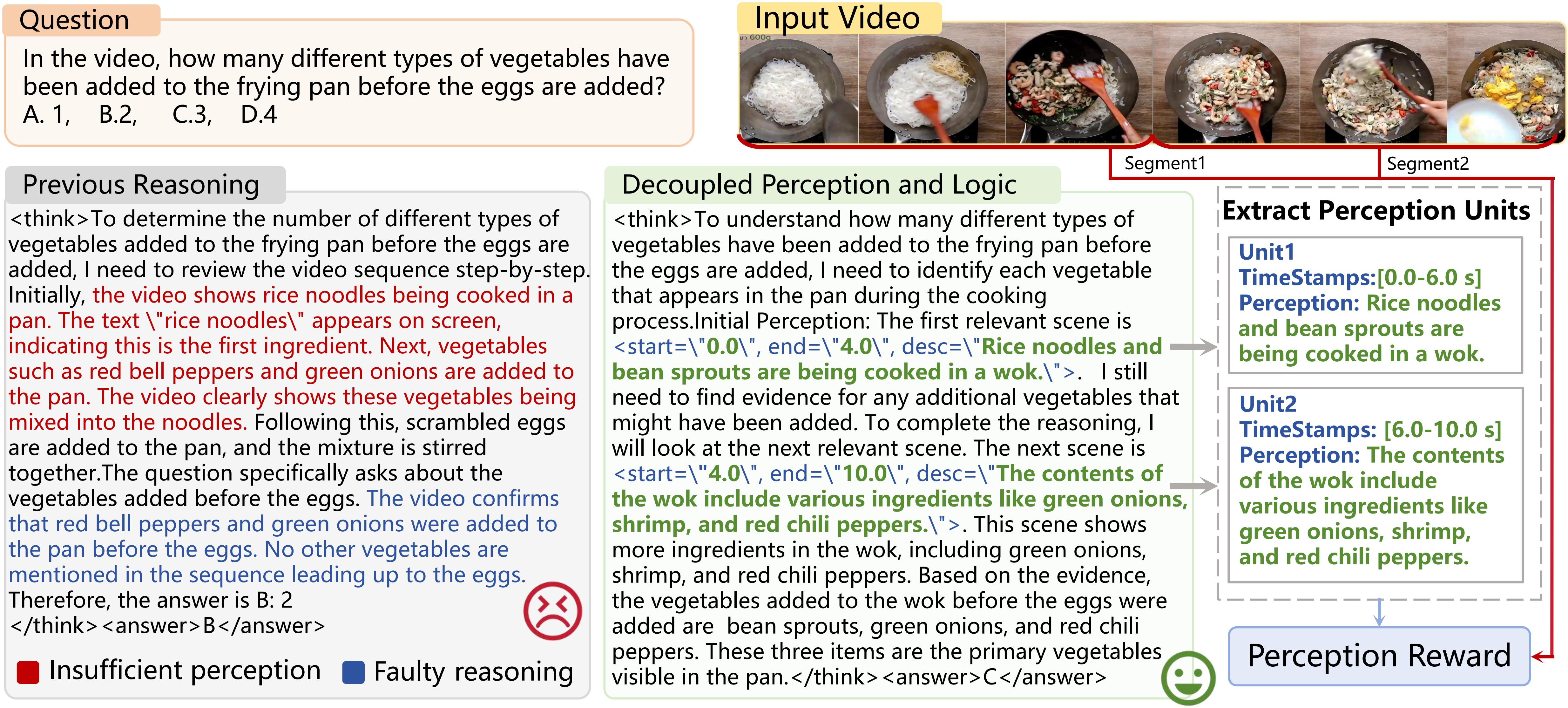}
        \caption{Overview of our framework. Given a video and a question, we first produce timestamped, segment-level evidence units that separate perception from reasoning. We score each unit with a factuality evaluator and optimize answer evidence consistency, yielding a video and question-grounded process reward for post-training.}
    \label{fig:intro}
    \end{figure}

Video Large Language Models (VideoLLMs)~\cite{llava-video,liLLaVAOneVisionEasyVisual2024,wangInternVideo2ScalingFoundation2024a} have achieved strong performance on video understanding benchmarks. For more complex reasoning tasks, many VideoLLMs~\cite{guo2025deepseek,wang2025videorft,zhang2025tinyllava,feng2025video} generate intermediate text to organize visual evidence before answering. While helpful, intermediate text can be misleading if it is not faithful to the video. Thus, reliability depends on grounding both intermediate reasoning and final answers in verifiable video evidence~\cite{jiang2025videop2r,meng2025open}.

A major failure mode arises from insufficient video perception, where models produce visual evidence that is not supported by the video. Recent works attempt to improve reasoning reliability by strengthening perception~\cite{feng2025video,wang2025videorft,zhang2025tinyllava,park2025deepvideo}.  However, existing training signals are often indirect and coarse-grained, making it difficult to design reliable rewards.
In many approaches, models first describe the video and then perform reasoning based on the description. 
As shown in the \emph{Previous Reasoning} block of Fig.~\ref{fig:intro}, the video description overlooks the segment where bean sprouts are added. Also, this reasoning is inaccurate and exhibits hallucinations. While humans can easily distinguish perception from reasoning, these two types of content are often mixed within the same textual span. As a result, reward signals can be affected by irrelevant reasoning noise. This leads to a practical question: \textbf{how should we design a reliable process reward for intermediate perceptual evidence?} We identify two key challenges:
(1) Perceptual content is difficult to isolate from reasoning text, making reward signals indirect.
(2) A video contains multiple events, yet evaluating a holistic description of the entire video is coarse-grained and unreliable.

Reliable supervision requires separating perception from reasoning and verifying perceptual evidence at a finer temporal granularity. This allows the evidence to be independently validated against the video. Therefore, we introduce  \textbf{Decoupled Perception and Logic (DPL)} and the \textbf{Perception Reward} built on top of it. DPL requires the model to first extract perception as a list of timestamped, segment-level evidence units, and then perform logical reasoning and produce the final answer. As shown in the \emph{Extract Perception Units} part of Fig.~\ref{fig:intro}, each unit includes a start time, an end time, and an evidence description.

Based on DPL, we introduce a {Perception Reward} that provides process supervision from two complementary perspectives.  First, we evaluate the alignment between each evidence unit and the video using a \textbf{Factual-Aware Evaluator (FAE)}, which measures whether the described event is supported by the corresponding video segment. 
Second, we introduce an answer–evidence consistency signal that verifies whether the final answer can be inferred from the extracted evidence units alone, without access to the video.
Together, these signals ensure that perceptual evidence is grounded in the video and that final answers are supported by the extracted evidence.
To train the evaluator, we construct \textbf{ANetHallu-117K}, a hallucination-aware dataset derived from ActivityNet~\cite{caba2015activitynet}. 
The dataset contains minimally different supported and hallucinated caption pairs, enabling the evaluator to learn fine-grained factual distinctions.

Experiments show that reliable process rewards improve both post-training performance and data efficiency across diverse video reasoning tasks. We train with 7k mixed samples across five task types. Notably, adding 0.5K video temporal grounding samples enables Video-DPL to outperform post-training baselines trained with over $5\times$ more data. Overall, Video-DPL demonstrates that reliable process rewards can significantly improve post-training performance in video reasoning.

    The contributions of this paper are:
    \begin{itemize}
    \item \textbf{Video-DPL}: We propose Video-DPL, a decoupled perception and logic framework that produces timestamped evidence units, enabling independently verifiable video grounding.
    \item \textbf{Perception Reward}: A segment-level process reward for post-training, encouraging truthful perception and evidence-grounded answering.
    \item \textbf{FAE}: A perception-focused factuality evaluator that scores evidence units, trained on minimally different supported and hallucinated caption pairs with a bias-aware pipeline.
    \end{itemize}

\section{Related Work}
\label{sec:related}

\subsection{Multimodal Perception Reasoning}
Recent multimodal reasoning methods improve reliability by strengthening perception and disentangling it from reasoning, e.g., Visionary-R1 and Reflection-V~\cite{xia2025visionary,jian2025look}. Recent video models have also moved in this direction. In video reasoning, Chain-of-Frames~\cite{ghazanfari2025chain} optimizes reasoning trajectories by describing key frames. Open-o3~\cite{meng2025open} emphasizes perception for video object localization. VIDEOP2R~\cite{jiang2025videop2r} follows a similar philosophy to Visionary-R1. 
However, existing methods' supervision and evaluation are often single-frame or whole-video, which can be noisy for dynamic videos; we instead target segment-level, verifiable perceptual evidence to support reliable process rewards.

\subsection{Reasoning VideoLLMs}
Recent work adapts RL-style~\cite{guo2025deepseek} post-training to VideoLLMs.  VideoR1~\cite{feng2025video} and DeepVideo-R1~\cite{park2025deepvideo} improve perception by perturbing the video input as a negative sample. VideoRFT~\cite{wang2025videorft} utilizes SigLip~\cite{tschannen2025siglip} during the perception stage to mitigate hallucinations.  TimeR1~\cite{wang2025time} and VideoChatR1~\cite{li2025videochat} place a greater emphasis on temporal perception and often lack consideration for general-purpose tasks.
While Rewatch-R1~\cite{zhang2025rewatch} proposes better cold-start data, its Observation reward is constrained by the expressive gap between finite textual descriptions and the model's nuanced attention patterns, leading to systematic evaluation bias. ARC-HunyuanVideo~\cite{ge2025archunyuanvideo7bstructuredvideocomprehension} can generate dense video descriptions with timestamps, but lacks the ability to reason within these segments, and does not address the hallucination problem. 
Current video reasoning training lacks research on process rewards for video perception. We identify the problem in the reasoning process and decouple perception and logic, providing reliable reward signals.

\subsection{Video Hallucination}
Longer reasoning chains reduce attention to visual input, thereby leading to hallucinations~\cite{liu2025more}. Consequently, hallucination control is crucial for video models. VideoRFT~\cite{wang2025videorft} utilizes SigLip~\cite{tschannen2025siglip} for visual-text consistency. Nevertheless, SigLip is not optimally suited for video. Utilizing a VideoLLM as a judge requires the model itself to be low-hallucinating and unbiased. However, the current models cannot achieve this. VidHalluc~\cite{li2025vidhalluc} evaluates hallucinations from three key dimensions, while VideoHallucer~\cite{wang2024videohallucer} focuses on the bias of judging the correctness of descriptions. To address this, we provide a controllable data generation pipeline for hallucination pairs and introduce a video hallucination judgement model. Our primary objective is to supply a reliable anti-hallucination reward.

\section{Factual-Aware Evaluator}
\begin{figure*}[t]
    \includegraphics[width=1\linewidth]{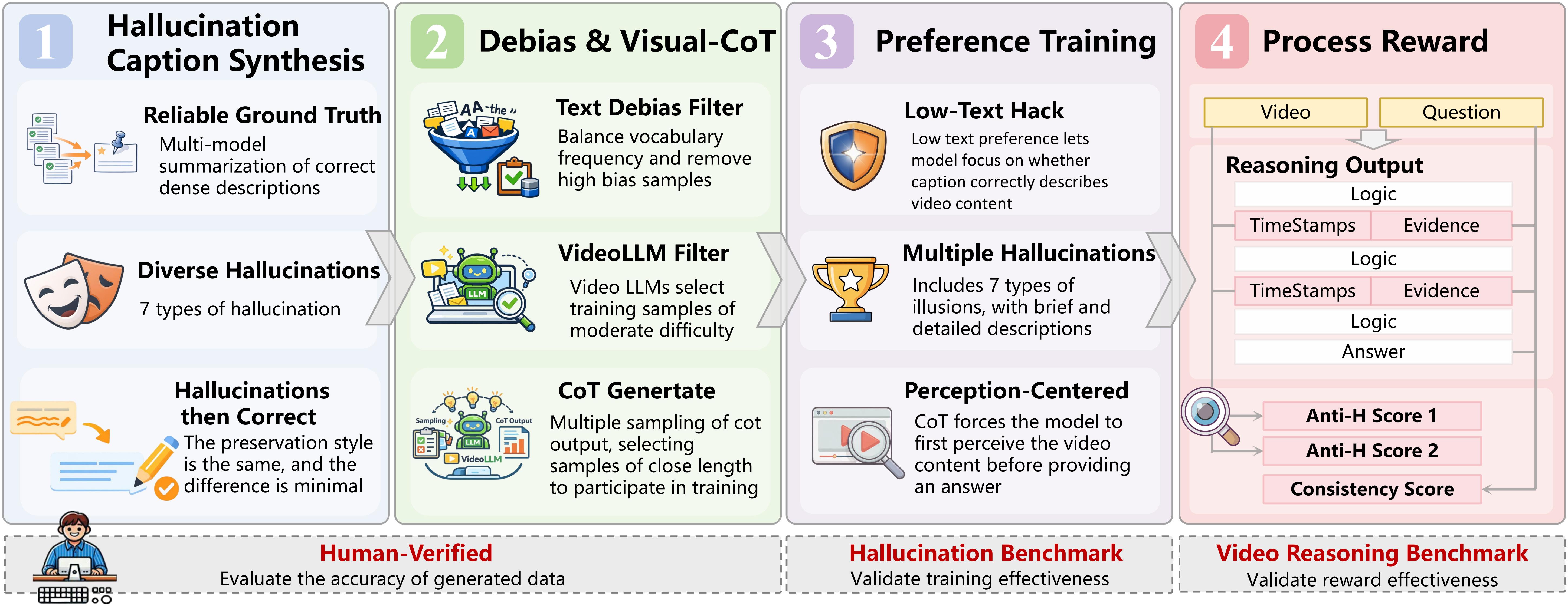}
    \caption{\textbf{FAE overview.} We synthesize matched-style hallucination caption pairs, apply text-debiasing and verification, and train a perception-centered judge with preference optimization. The judge’s support confidence for a caption given a video segment is later used as an anti-hallucination reward signal for post-training.}
\label{fig:method}

\end{figure*}

A key challenge in post-training is to derive an anti-hallucination reward that reflects whether a candidate statement is supported by the \emph{visual content} of a specific video segment, rather than by linguistic plausibility. We propose the \textbf{Factual-Aware Evaluator (FAE)}, a perception-centered judge that takes a segment $V_{[t_1,t_2]}$ and a candidate caption $C$, and outputs a support decision together with its confidence. We use this support confidence as the factuality score and later convert it into an anti-hallucination reward (see Sec~\ref{sec:reward}). We train FAE on AnetHallu-117K, built from ActivityNet (ANet)~\cite{caba2015activitynet}, using matched-style hallucination pair construction and preference optimization to suppress language shortcuts, as shown in the pipeline in the Fig~\ref{fig:method}. 

Furthermore, we use the following notation throughout this section: the segment-level ground-truth caption is $C_{gt}$; a minimally different caption pair is $(C_{pos}, C_{neg})$; and a preference instance is $(x,y_w,y_l)$, where $x$ is the input context and $y_w$/$y_l$ are the preferred and dispreferred outputs.

\subsection{Re-annotate Dense Captions}
ANet covers diverse real-world scenes, but its original annotations are coarse and cannot directly support subtle hallucination-pair construction. We therefore re-annotate each segment and produce high-fidelity segment-level ground-truth captions $C_{gt}$ for $V_{[t_1,t_2]}$. Concretely, we first sample candidate dense descriptions from multiple models and then aggregate them with a stronger closed-source LLM to improve factual completeness and consistency. All subsequent positive/negative caption pairs are anchored to the same $C_{gt}$ to preserve semantic alignment and language comparability. 

\subsection{Low-bias generation via Hallucinate-then-Correct}
Our goal is to construct caption pairs whose labels are difficult to predict from surface text patterns, such as length, fluency, and template cues, without looking at the video. To this end, we adopt a two-stage \textbf{Hallucinate-then-Correct} pipeline that keeps style and length consistent within each pair.

We cover five hallucination types to improve robustness: (1) \emph{Attribute Modification}, (2) \emph{Quantity Modification}, (3) \emph{Action Substitution}, (4) \emph{Detail Conflation}, and (5) \emph{Temporal Reordering}. Given $C_{gt}$ and a chosen type, we first generate a negative caption $C_{neg}$ that introduces a targeted factual error. In the correction stage, we condition on both $C_{neg}$ and $C_{gt}$ and generate a positive caption $C_{pos}$ that fixes only the targeted error while preserving surrounding phrasing.  This process yields raw preference pairs with a closely matched writing style.

\begin{figure}[t]
    \centering
        \centering
        \begin{minipage}{0.32\linewidth}
            \centering
            \includegraphics[width=\linewidth]{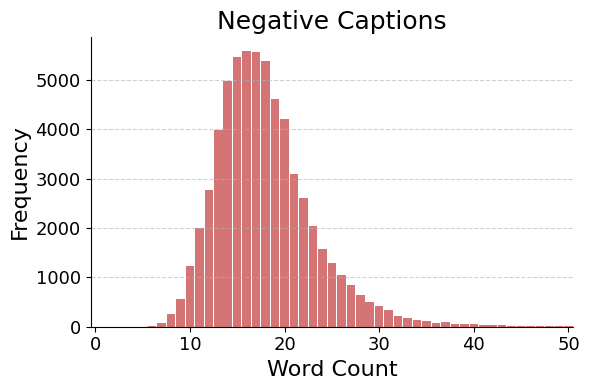} 
            \subcaption{Negative Captions} 
        \end{minipage}%
        \hfill 
        \begin{minipage}{0.32\linewidth}
            \centering
            \includegraphics[width=\linewidth]{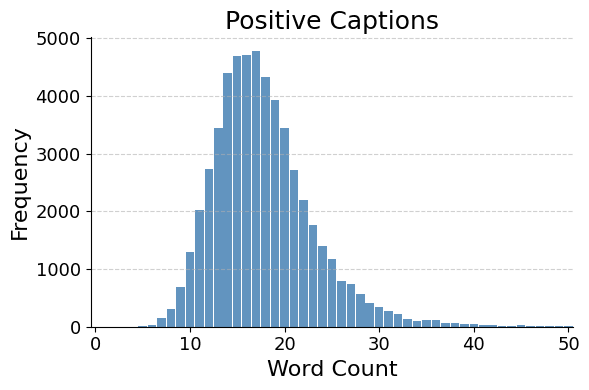} 
            \subcaption{Positive Captions} 
        \end{minipage}
         \begin{minipage}{0.35\linewidth}
        \centering
        \includegraphics[width=\linewidth]{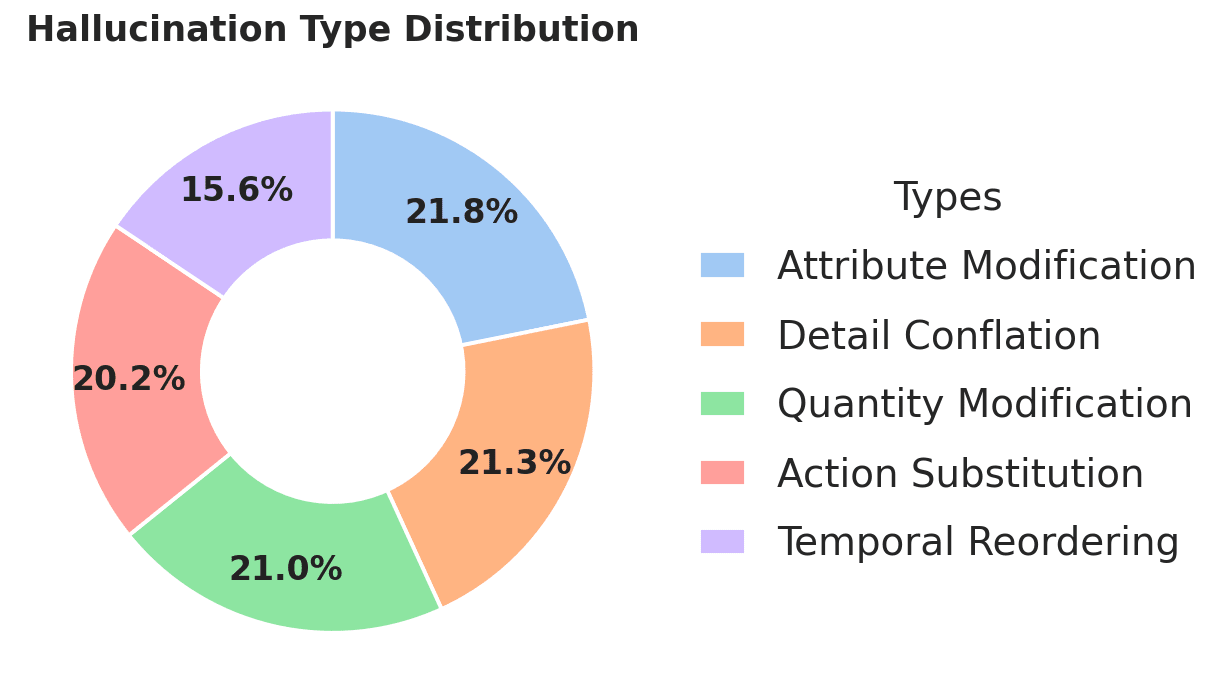} 
        \subcaption{Distribution of each type of hallucination}
        \label{fig:hallu_type}
        \end{minipage}
        \caption{Statistical data for ANetHallu-117K, covering its length distribution and type distribution.} %
        \label{fig:hallu}
\end{figure}

\begin{table}[t]
\centering
\begin{minipage}[t]{0.49\linewidth}
\centering
\caption{Hallucination data statistics by response format.}
\label{tab:hallu_data_breakdown}
\setlength{\tabcolsep}{9pt}
\begin{tabular}{@{}lccc@{}}
\toprule
\textbf{Type} & \textbf{Yes} & \textbf{No} & \textbf{Total} \\
\midrule
Direct & 34{,}621 & 42{,}618 & 77{,}239 \\
CoT  & 19{,}963 & 19{,}969 & 39{,}932 \\
\midrule
{Total} & {54{,}584} & {62{,}587} & {117{,}171} \\
\bottomrule
\end{tabular}
\end{minipage}

\hfill
\centering
%
%
\begin{minipage}[t]{0.49\linewidth}
\centering

\caption{Human agreement on hallucination labels.}
\label{tab:human_agreement_hallu}
\begin{tabular}{@{}lccc@{}}
\toprule
\textbf{Split} & \textbf{Sampled} & \textbf{Matched} & \textbf{Score(\%)} \\
\midrule
Yes & 200 & 195 & 97.5 \\
No  & 200 & 156 & 78.0 \\
\midrule
{Overall} & 400 & 349 & 87.3 \\
\bottomrule
\end{tabular}
\end{minipage}

\end{table}

\subsection{Quality Filtering and Verification}
Although caption-pair styles are aligned, the data generated by LLM may still contain fixed preferences. A model may exploit these preferences instead of video evidence. To reduce such text shortcuts, we perform iterative \textbf{Text Bias Elimination} based on normalized word-frequency statistics computed after standard text normalization. Specifically, we compute relative frequency ratios over the positive/negative sets and build biased vocabularies $\mathcal{W}_{pos}$ and $\mathcal{W}_{neg}$ from top-ranked words.
\begin{align}
    R_{neg}(w) = \frac{f_{neg}(w)}{\max(f_{pos}(w),1)},
\end{align}
where $f(\cdot)$ is word frequency in the corresponding set. We then assign a normalized penalty
\begin{align}
    P_{neg}(w) = \frac{R_{neg}(w)}{\max_{w' \in \mathcal{W}_{neg}} R_{neg}(w')},
\end{align}
and compute a caption-level bias score
\begin{align}
    S(c_{neg}) = \sum_{w \in W(c_{neg}) \cap \mathcal{W}_{neg}} P_{neg}(w),
\end{align}
where $W(c_{neg})$ denotes words in $c_{neg}$. We apply the same procedure to positive captions, run filtering for 15 iterations, and remove the top 2\% biased samples per iteration with vocabulary size $N=30$.
As shown in Fig.~\ref{fig:hallu} and Table~\ref{tab:hallu_data_breakdown}, the resulting positive or negative captions have similar length and balanced format coverage, reducing superficial style leakage.

After automatic debiasing, we run double-annotator verification on a held-out subset. The agreement pattern in Table~\ref{tab:human_agreement_hallu} indicates that hallucination-labeled cases remain harder than non-hallucination ones, while overall consistency is still high. This suggests we are retaining challenging negatives.

\subsection{CoT-Enhanced Perception for Judging}
Word-level filtering alone may be insufficient because higher-order linguistic cues can remain. We therefore introduce \textbf{CoT-Enhanced Perception}: before outputting the final support label, the judge first produces a short perception summary grounded in the video segment.

For each video-caption input $(V_{[t_1,t_2]}, C_{test})$, we prompt the model to (i) generate a factual perception summary $R_{desc}$ for the segment and then (ii) judge whether $C_{test}$ is supported. The final tuple is
\begin{align}
    (V_{[t_1,t_2]}, C_{test}, R_{desc}, A),
\end{align}
where $A$ is a binary label. To avoid degenerate preference signals, we discard instances where the judge always predicts the same label across multiple decoding, and keep instances with non-trivial uncertainty as preference data.
We then optimize the judge on a hybrid preference dataset using ORPO, which improves both label accuracy and perception-grounded reasoning.

\subsection{Hybrid Preference Training with ORPO}
We train FAE with a hybrid preference dataset $\mathcal{D}$ containing pairs $(x,y_w,y_l)$, where $x$ is a video-caption input. The first pair type emphasizes \emph{Answer Accuracy} (direct vs. direct) for efficient prediction. The second emphasizes \emph{Reasoning Accuracy} (CoT vs. CoT) to enforce perception-grounded judgment. This mixture improves both robustness and efficiency.

We optimize with ORPO~\cite{hong2024orpo}.
This training yields a factual evaluator that is less sensitive to superficial text cues and more reliant on segment-level visual evidence. 

\section{Post-Training of the Perception Process Rewards}
This section presents a post-training framework with process rewards. As shown in the rightmost module of Fig.~\ref{fig:method}, we build the perception reward on top of the Decoupled Perception and Logic framework, incorporating an anti-hallucination signal from FAE and a consistency signal.
Our goal is to make the visual evidence reliable and the answers based on it.

\subsection{Decoupled Perception and Logic}
In video reasoning, free-form CoT often mixes visual perception and logical reasoning in the same text span. This coupled expression makes the intermediate process hard to align, which in turn makes it difficult to provide effective process reward signals. To address this issue, we explicitly decouple perception from logic. Specifically, we require the model to first output structured perception evidence units, where each unit contains a start time, an end time, and an evidence description, in the following format:
\begin{align*}
     \texttt{<start="t1",end="t2",desc="evidence description">}.
\end{align*}

The remaining content is treated as logical reasoning and final answer generation. This decoupling brings two direct benefits. First, perception is converted from holistic descriptions into timestamped evidence units, so each claim can be verified independently on its corresponding video segment, reducing noise and ambiguity in long-video evaluation. Second, process supervision becomes functionally decomposed and more stable, one reward focuses on evidence factuality, while another enforces consistency and traceability between evidence and the final answer. As a result, process rewards become more computable, attributable, and scalable.

\subsection{Perception Reward}
\label{sec:reward}
Based on the decoupled output structure, we designed two types of rewards to jointly enhance perceptual accuracy and problem relevance.

\textbf{Anti-Hallucination.}
Each generated answer contains a set of perception units, where each unit consists of a temporal span and a textual description.
For an perception unit $p_i = (t_i^{start}, t_i^{end}, d_i)$, we extract the corresponding video clip and feed it, together with the description $d_i$, into the FAE.
The FAE outputs the probabilities that the description is supported ($P_Y$) or not supported ($P_N$) by the clip.
We normalize the support probability as:
\begin{align}
p_i = \frac{P_Y}{P_Y + P_N}.
\end{align}

Thus, our anti-hallucination reward ($R_{ha}$) is defined as a weighted sum of all perception units:
\begin{align}
R_{ah}(y) = \frac{1}{max(0.6+0.8\times |\mathcal{E}(y)|,|\mathcal{E}(y)|)} \sum_{p_i \in \mathcal{E}(y)} p_i,
\end{align}
where $\mathcal{E}(y)$ is the perception list, and $|\mathcal{E}(y)|$ is defined as the number of perception.

This reward directly penalizes evidence descriptions that appear plausible but are not grounded in the corresponding video segment.

\textbf{Consistency.}
We measure answer traceability by testing whether the final answer is recoverable from the extracted perception alone.
Concretely, we feed the question $q$ and the perception list $\mathcal{E}(y)$ without video input to a reference model and obtain:
\begin{align}
\hat{a}_{ref} &= \text{ref\_model}(q,\mathcal{E}(y)).
\end{align}
We then compare it with the policy model's final answer $\hat{a}$:
\begin{align}
R_{cons}(y) &= \text{Match}(\hat{a}_{ref},\hat{a}).
\end{align}
Here $\text{Match}(\cdot,\cdot)$ is task-dependent.
Intuitively, when the extracted perception is sufficient and factual, the same answer should be reachable from $\mathcal{E}(y)$ alone; low consistency typically indicates missing perception or a disconnect between the perception and the answer.

Together with $R_{ah}$, these rewards are complementary: $R_{ah}$ constrains perception factuality, while $R_{cons}$ enforces evidence-grounded answering.

\subsection{Cold-Start Data Construction}
\label{sec:cold_start}
\begin{figure}[t]
  \centering
    \begin{minipage}{0.32\linewidth}
        \centering
        \includegraphics[width=\linewidth]{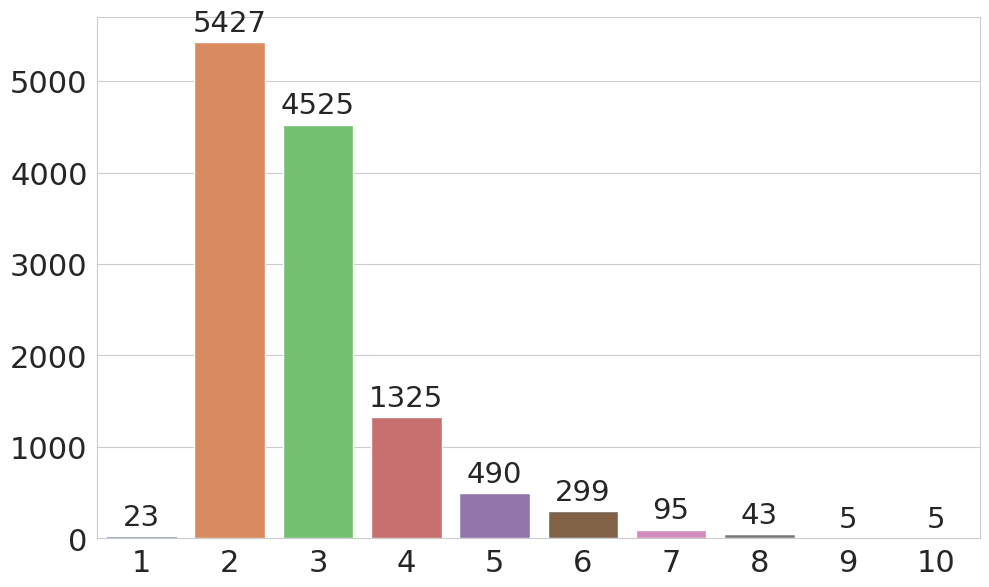} 
        \subcaption{Perception Count Distribution of Cold Start Data.} 
        \label{fig:event_count}
    \end{minipage}
    \begin{minipage}{0.32\linewidth}
        \centering
        \includegraphics[width=\linewidth]{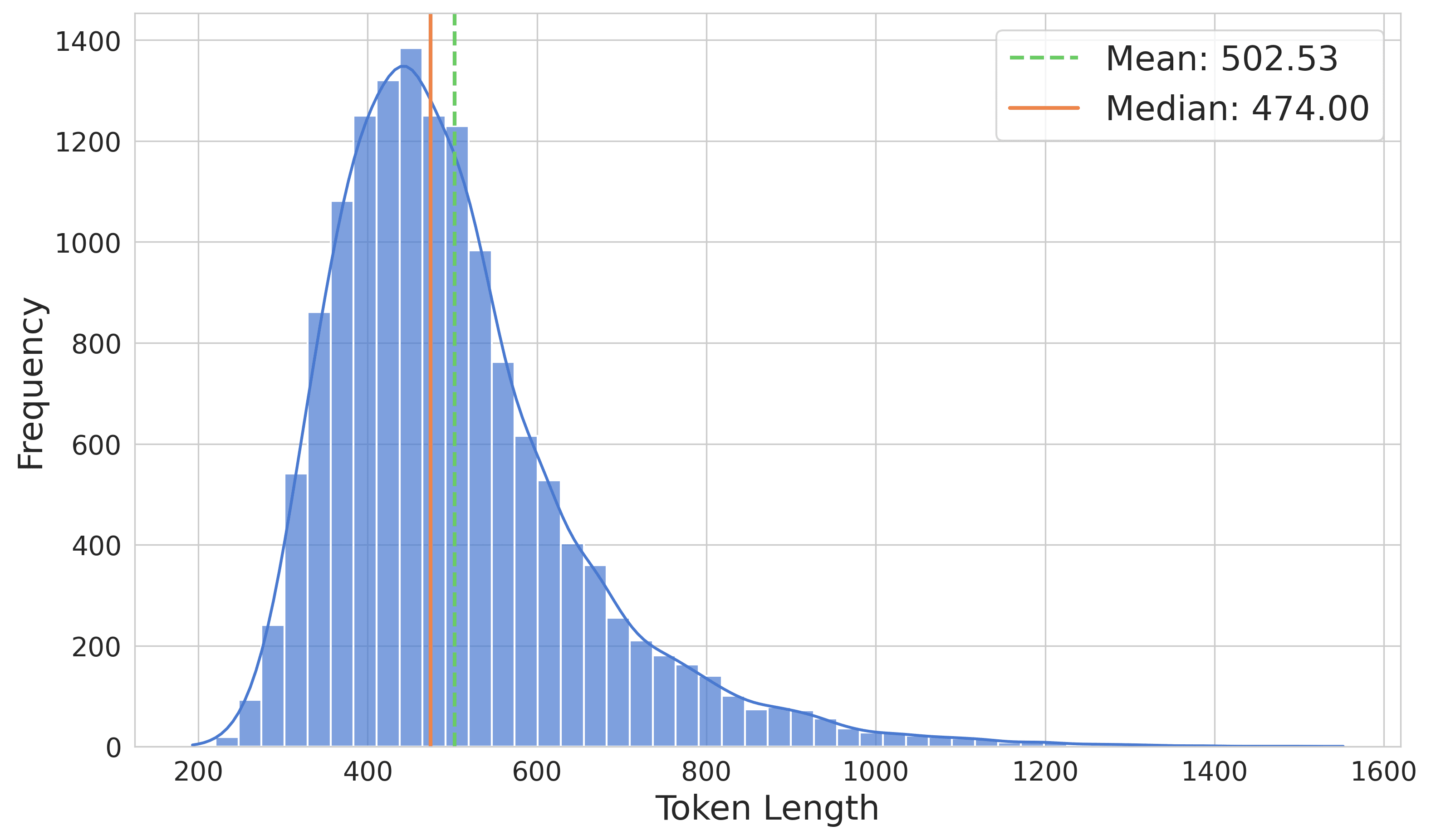} 
        \subcaption{Token Count Distribution of Cold Start Data.}
        \label{fig:reasoning_length}
    \end{minipage}
    \begin{minipage}{0.32\linewidth}
        \centering
        \includegraphics[width=\linewidth]{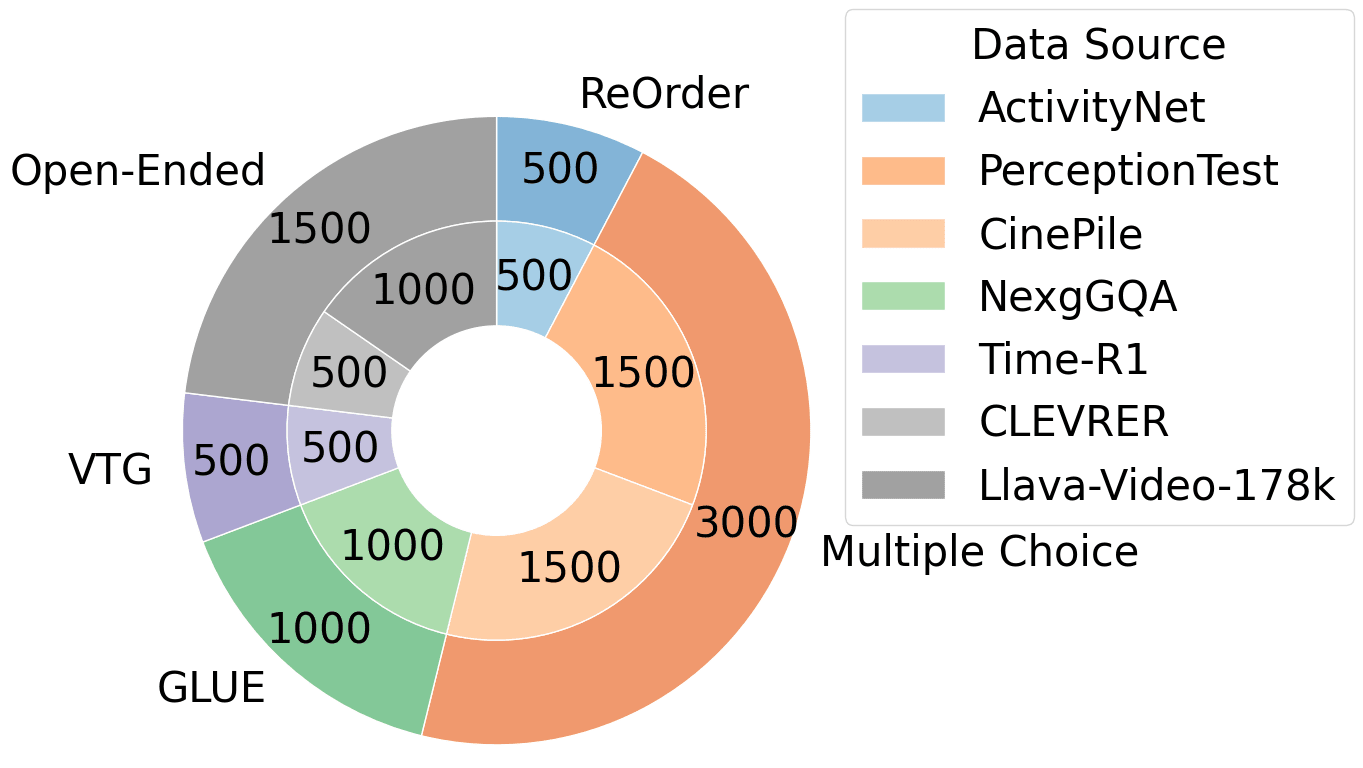} 
        \subcaption{The RL data type includes five task types.}
        \label{fig:rl_data}
    \end{minipage}
  
    \caption{The statistical distribution of data during the cold start and reinforcement learning phases}
    \label{fig:posttrain_stats}
\end{figure}

We build cold-start data to instantiate DPL style reasoning traces. The pipeline has three steps. First, we perform frame by frame annotation to obtain timestamp-aligned captions. Second, we feed the timestamped frame captions into an LLM to aggregate local observations into coherent event level evidence. Third, the LLM rewrites the aggregated evidence into CoT outputs that explicitly include DPL perception units, followed by logic and the final answer.
We apply this pipeline to open-ended QA and  multi-choise task based on NextQA~\cite{xiao2021next}. To enable the model to perceive video content in a non-temporal manner, we designed a reordering task based on ANet~\cite{caba2015activitynet}. This task requires the model to sequentially perceive and sort each event involved. We further use FAE filtering and remove samples with invalid format. The final cold-start set contains 14K high-quality samples, shown as Fig.~\ref{fig:event_count} and Fig.~\ref{fig:reasoning_length}.

\subsection{Training with GRPO}


The key to GRPO is the reward design. Our goal is to ensure the model responds accurately while minimizing hallucinations. The output of the model will be defined as $y$.
Our training data mainly contains five kinds of tasks: Video Temporal Grounding (VTG), Multiple Choice (MC), MC\&VTG (GLUE), ReOrdering, and Open-Ended (OE), illustrated in Fig.~\ref{fig:rl_data}. Except for the OE task, we use rule matching. And for the OE task, we use an open-source verification model (VM)~\cite{xVerify}. Given the question $Q$, the ground truth answer $G$, and the model's generated answer $A_{\text{gen}}$ as input, we obtain the token probabilities $P_{Ic}$ and $P_C$ for the Incorrect and Correct labels, respectively.
Thus, the accuracy reward $R_{acc}$ is following:
\begin{align}
R_{acc}(y) =
\begin{cases}
  IoU(G,\hat{t}), & \text{if VTG} \\
  Match(G,\hat{a}), & \text{if MC } \\
    \frac{P_C}{P_{Ic}+P_C},   & \text{if OE}\\
  Match(G,\hat{a})+IoU(G,\hat{t}), & \text{else} 
\end{cases}
\end{align}
where $\hat{a}$ and $\hat{t}$ denote the parsed final answer and temporal span from the model output $y$, respectively, and $Match$ is defined as choosing the same options or having the same sorting order.

We additionally include a format reward $R_f$ to supervise both the reasoning format and the evidence format, encouraging outputs with a consistent thinking structure and machine-parsable evidence units. We further incorporate an anti-hallucination reward $R_{ah}$ and a consistency reward $R_{cons}$.

The overall reward for each sample is:
\begin{align}
R(y) = R_{acc}(y) + 0.5R_f(y) + 0.2 \cdot (\mathbf{1}[R_{acc}(y) > 0.5 ]\cdot R_{ah}(y) + R_{cons}(y)).
\end{align}



\begin{table}[t]
\captionsetup{skip=2pt}              
\captionsetup[subtable]{skip=1pt}    
\centering
\caption{Results on hallucination benchmarks and reward analysis.
$^{\dagger}$ denotes rewriting questions as captions. Diff denotes the absolute factual, accuracy gap.
Red numbers indicate improvements over Qwen2.5VL. $\Delta = R_{yes} - R_{no}$.}
\label{tab:halluc_and_reward}

\begin{subtable}[t]{0.60\linewidth}
\centering
\subcaption{Hallucination benchmark results.}
\label{tab:hallu}
\small
\setlength{\tabcolsep}{2pt}
\resizebox{\linewidth}{!}{
\begin{tabular}{lccccc}
\toprule
\multirow{2}{*}{Model}
& \multicolumn{2}{c}{VideoHallucer~\cite{wang2024videohallucer}}
& \multicolumn{2}{c}{HEAVEN$^{\dagger}$~\cite{gao2025exploringhallucinationlargemultimodal}}
& VidHalluc~\cite{li2025vidhalluc} \\
\cmidrule(r{0.8em}){2-3} \cmidrule(r{0.8em}){4-5} \cmidrule(r{0.8em}){6-6}
& Acc$\uparrow$ & Diff$\downarrow$ & Acc$\uparrow$ & Diff$\downarrow$ & MC Acc$\uparrow$ \\
\midrule
GPT-4o~\cite{openai2024gpt4ocard} & 74.6 & \textbf{0.9} & - & - & \textbf{91.0} \\
Gemini-1.5-Pro~\cite{geminiteam2024gemini15unlockingmultimodal} & 62.9 & 41.3 & - & - & 79.2 \\
Qwen2.5VL~\cite{baiQwen25VLTechnicalReport2025} & 67.8 & 22.4 & 63.7 & 28.9 & 89.2 \\
\midrule
FAE (Ours)
& \textbf{75.4} \textcolor{red}{(+7.6)} & 7.3 \textcolor{red}{(-15.1)}
& \textbf{72.9} \textcolor{red}{(+9.2)} & \textbf{2.1} \textcolor{red}{(-26.8)}
& 90.3 \textcolor{red}{(+1.1)} \\
\bottomrule
\end{tabular}
}
\end{subtable}
\hfill
\begin{subtable}[t]{0.37\linewidth}
\centering
\subcaption{Reward Score on HAVEN$^{\dagger}$}
\label{tab:reward_boundary}
\small
\setlength{\tabcolsep}{4pt}
\resizebox{\linewidth}{!}{
\begin{tabular}{lccc}
\toprule
Model & $R_{yes}\uparrow$ & $R_{no}\downarrow$ & $\Delta\uparrow$ \\
\midrule
SigLip~\cite{tschannen2025siglip} & 40.96 & 29.64 & 11.32 \\
Qwen2.5VL~\cite{baiQwen25VLTechnicalReport2025} & 54.34 & 33.07 & 21.27 \\
{FAE (Ours)} & \textbf{65.44} & 35.82 & \textbf{29.62} \\
\bottomrule
\end{tabular}
}
\end{subtable}

\end{table}
\section{Experiment}
\subsection{Setups}
\textbf{Training Details}
For the Factual-Aware Evaluator, we use LoRA~\cite{hu2022lora} based on Qwen2.5VL 7B, training for one epoch with a learning rate of 1e-4, where $\lambda$ in ORPO is 0.5, with a maximum of 16 frames and the total number of pixels limited to $1200\times28\times28$.
For the reasoning model, training includes a cold-start SFT and RL using GRPO. The cold-start data incorporates 20K image CoT data from the Reason-RFT-CoT-Dataset~\cite{tan2025reason} and our VideoDPL-14K. We sample  7k questions for training, sourced from LLava-Video-178K~\cite{llava-video}, CLEVRER~\cite{yi2020clevrer}, Time-R1~\cite{wang2025time}, Cinepile~\cite{rawal2024cinepilelongvideoquestion}, NextGQA~\cite{xiao2024can}, and ANet~\cite{caba2015activitynet}, as show in Fig.~\ref{fig:rl_data}. For both SFT and RL, the learning rate is set to 1e-6. We set the number of generations to 8, and the batch size to 7; the $\beta$ in GRPO is set to 0.01. The maximum number of frames and total pixels are set to $64$ and $2048\times 28\times 28$. Our training code is optimized based on ms-swift~\cite{zhao2024swiftascalablelightweightinfrastructure} to adapt to the anti-hallucination reward, which incurs significant computational latency. Thus, we implement GRPO training without additional time consumption.  \\
\textbf{Evaluation Details}
First, we evaluate the Factual-Aware Evaluator, including the VidHalluc~\cite{li2025vidhalluc}, VideoHallucer~\cite{wang2024videohallucer}, and HEAVEN~\cite{gao2025exploringhallucinationlargemultimodal}. We set the maximum number of frames to 16, the number of pixels to $2400\times 28\times 28$.
For video reasoning models, the evaluation benchmarks include three types: reasoning, general, and temporal. The reasoning benchmarks include: Video-Holmes~\cite{cheng2025videoholmesmllmthinklike}, MMVU~\cite{zhao2025mmvu}, VCR-benchmark~\cite{qi2025vcrbenchcomprehensiveevaluationframework},Video-tt~\cite{zhang2025towards}, VSI-bench~\cite{yang2025thinking}. The general benchmark is VideoMME~\cite{fu2025video}, and the temporal benchmark is TVG-Bench~\cite{wang2025time}. We set the maximum number of frames to 64, the number of pixels to $6000\times 28\times 28$.

\begin{table*}[t]
\centering
\caption{The results of Video Reasoning and General Benchmarks. For VCR-Bench, Video-tt,  and TVG-Bench, we re-tested the performance metrics of each model. The metric of TVG-Bench is R1@0.5. Considering the fairness, for Videorchat-R1.5 and Rewatch-R1, we limited the maximum number of frames. The other models were evaluated using the default settings. In the Training Data column, we compared the data scales in the cold start and RL phases, in the form of cold start + RL phase.}
\label{tab:main}
\small
\setlength{\tabcolsep}{3.5pt}
\resizebox{\linewidth}{!}{
\begin{tabular}{lccccccccc}
\toprule
\multirow{2}{*}{Model} &\multirow{2}[2]{*}{\makecell{Training \\ Data}$\downarrow$} & \multicolumn{5}{c}{ Reasoning } & \multicolumn{1}{c}{ General }& \multicolumn{1}{c}{ Temporal }&\multirow{2}{*}{Avg} \\

\cmidrule(r{1em}){3-7} \cmidrule(l{1em}){8-8}\cmidrule(l{1em}){9-9}

& & {Vid-Holmes} & {MMVU(mc)} & {VCR}& {VSI} & {Vid-tt(mc)} & {VideoMME}  & {TVG} &\\

\midrule

\rowcolor{gray!25}
\multicolumn{3}{l}{Closed-Source Model} & & & &&& & \\
GPT-4o~\cite{openai2024gpt4ocard} & -& 42.0 & 67.4 & 46.9 & 34.0 & 46.6 & 71.9& - & -\\
Gemini-2.0-Flash~\cite{comanici2025gemini} & -& 30.6 & 65.9 & 51.7 & 45.4 & - & -& -& - \\

\midrule

\rowcolor{gray!25} 
\multicolumn{3}{l}{3B Open-Source Model} & & & && && \\
Qwen2.5VL~\cite{baiQwen25VLTechnicalReport2025} & - & 17.0 & 53.0 & 24.8 & 28.6 & 36.9 & 52.1 &14.3&32.4 \\
Time-R1~\cite{wang2025time}  & 0+2.5K& 32.2 & 52.5 & 22.9 & 28.5  & 34.9 & 53.4&\underline{21.0}&35.3  \\
Video-RFT~\cite{wang2025videorft} & 102+8K & \underline{34.3} & \underline{52.6} & \underline{25.8} & \underline{31.9}  & \underline{35.2} & \underline{54.7}&10.6 &\underline{34.7} \\
Video-DPL (our) & 34+7K  & \textbf{39.9} & \textbf{59.2} & \textbf{29.4} & \textbf{34.1} & \textbf{38.9}  & \textbf{57.3}&\textbf{24.9}&\textbf{40.5} \\
\midrule

\rowcolor{gray!25}
\multicolumn{3}{l}{7B Open-Source Model}  & && & & & &\\
Qwen2.5VL~\cite{baiQwen25VLTechnicalReport2025}  & -& 27.8 & 63.3 & 29.7 & 31.8 & 39.9 & 60.1&18.7& 38.8  \\
Time-R1~\cite{wang2025time}  & 0+2.5K& 37.1 & 64.2 & 29.6 & 28.4 & 39.0  & 60.9&29.6& 41.3 \\
Video-R1~\cite{feng2025video} & 165+8K& 36.5 & 64.2 & {31.3} & 35.8 & 39.1 & 59.3&13.7& 40.0 \\
Video-RFT~\cite{wang2025videorft}  & 102+8K& {39.8} & \textbf{68.5} & 30.6 & 36.8 & 40.3  & 59.8 &15.4&41.6 \\
Videochat-R1~\cite{li2025videochat}  & 0+18K& 33.0 & 63.8 & 30.0 & 31.3 & 39.1 & 60.8&\underline{30.1}&41.2  \\
Videochat-R1.5~\cite{yan2025videochat}  &0+80K& 38.5 & 64.5 & 30.0 & \textbf{40.6} & {41.3} & \underline{61.2}&26.8& \underline{43.2} \\
Rewatch-R1~\cite{zhang2025rewatch}  &315+40K& \underline{41.0} & 65.1 & \textbf{35.9} & {31.5} & \textbf{44.0} & {60.1}&14.8& {41.9} \\

Video-DPL (our) & 34+7K & \textbf{41.6} & \underline{65.9} & \underline{33.9} & \underline{38.5} &\underline{42.7} & \textbf{61.3} &\textbf{31.7}&\textbf{45.1} \\

\bottomrule
\end{tabular}
}

  \vspace{-4mm}
\end{table*}

\subsection{Hallucination Evaluation}
\textbf{Benchmark Improvement:}
We first evaluate hallucination detection performance, as shown in Tab.~\ref{tab:hallu}. We use VideoHallucer~\cite{wang2024videohallucer} for binary judgment, the open-source HEAVEN~\cite{gao2025exploringhallucinationlargemultimodal} reformatted as caption classification, and the MCQ task from VidHalluc~\cite{li2025vidhalluc}.
FAE significantly improves hallucination detection accuracy.
On VideoHallucer, it surpasses GPT-4o and improves the base model by 7.5\% and 9.2\%.  More importantly, FAE substantially reduces prediction bias toward either 'Y' or 'N'. It narrows the accuracy gap between correct and incorrect classifications by 15.1\% and 26.8\%. This suggests that our data construction strategy improves prediction fairness and encourages the model to rely on visual evidence rather than dataset priors. Despite being trained exclusively on binary classification tasks, our model demonstrates strong generalization to the MCQ task. 
This indicates that hallucination detection learned through factual consistency evaluation transfers across task formats. 

\textbf{Stronger Distinction:}
We further analyze the reward discrimination ability in Tab.~\ref{tab:reward_boundary}. 
FAE produces a substantially larger reward gap between factual ($R_{yes}$) and hallucinated ($R_{no}$) samples. 
Compared with SigLip and the base Qwen2.5VL, FAE increases the reward margin $\Delta$ to 29.62, indicating a clearer separation between correct and hallucinated evidence. 
Such a sharper reward boundary provides a more reliable supervision signal for reinforcement learning, enabling the model to better distinguish valid visual evidence from hallucinated content.

\subsection{Main Results}
We compare the models that were post-trained based on Qwen2.5VL. These baselines have different focuses in training.
Video-DPL achieves consistent improvements across both 3B and 7B scales. 

\textbf{SOTA Results:}
Overall, our Video-DPL achieves a significant average improvement in both the 3B and 7B parameter scales, as shown in Tab.~\ref{tab:main}.
The gains are particularly pronounced for 3B models. This suggests that improving perception reliability is especially beneficial for weaker models.
At the 7B scale, Video-DPL also achieves consistent improvements. 
On Video-Holmes, it outperforms all open-source models, and it achieves the second-best performance on several other reasoning benchmarks.
Notably, although our model is not trained with explicit object localization supervision, it still achieves strong performance on VSI-Bench. 
This result suggests that improving perception reliability can transfer to spatial reasoning tasks without dedicated localization supervision. 

\textbf{Data Efficient:}
While achieving the best performance, Video-DPL demonstrates excellent data efficiency. In the comparison at the 7B scale, our model uses only 34K samples for cold-start training and 7K samples for RL. Except for Time-R1, all other models require substantially more training data. For example, Rewatch-R1 uses about $7\times$ more training data than our method. It is noteworthy that although we only used 0.5K VTG training data, the performance has already surpassed that of TimeR1 trained with 2.5K data over 5 epochs in TVG-Bench.

\subsection{Ablation Study}

\noindent\textbf{Hallucination Training Data} 

As shown in Tab.~\ref{abl:hallu}, the mixed training data helps the model achieve better performance in hallucination evaluation. 
Specifically, we provide the accuracy difference for the binary classification task. The addition of CoT data has significantly alleviated the classification bias. 
This proves the effectiveness of the CoT data design, forcing the model to answer based on the video content. Additionally, directly answering `Yes' or `No' is  demonstrably more efficient than answering with CoT.

\begin{wraptable}{r}{0.48\linewidth}
\vspace{-6mm}
\centering
\caption{Ablation study on the training data. It can be found that the combination of CoT and direct output can significantly enhance the hallucination evaluation ability.}
\label{abl:hallu}
\resizebox{0.95\linewidth}{!}{
\setlength{\tabcolsep}{4pt}
\begin{tabular}{ll|ccccc}
\toprule
\multicolumn{2}{c}{Traing Set} & \multicolumn{2}{c}{HAVEN$^{\dagger}$ ~\cite{gao2025exploringhallucinationlargemultimodal}} & \multicolumn{3}{c}{VideoHallucer~\cite{wang2024videohallucer}} \\
\cmidrule(lr){3-4} \cmidrule(lr){5-7}
Direct & CoT & Acc$\uparrow$ & Diff$\downarrow$ & Acc$\uparrow$ & Diff$\downarrow$ & Gather$\uparrow$ \\
\midrule
& & 63.7 & 28.9 & 67.8 & 22.4 & 41.7\\
& $\checkmark$ & \underline{72.6} & \textbf{0.6} & 71.6 & \textbf{6.3} & \underline{51.6} \\
$\checkmark$ & & 70.1 & 4.6 & \underline{72.5} & 28.8 & 47.5  \\
$\checkmark$ & $\checkmark$ & \textbf{72.9} & \underline{2.1} & \textbf{75.3} & \underline{7.3} & \textbf{54.2}  \\
\bottomrule
\end{tabular}
}
\vspace{-4mm}
\end{wraptable}

The performance of the model trained with a mixture is also better than that trained only with CoT. The \emph{Gather} column shows the comprehensive performance, which requires both answers to paired questions to be correct. Obviously, mixed training is superior to other solutions. Thus, hybrid training yields more efficient responses and more accurate hallucination judgments.

\begin{figure}[t]
  \centering
    \begin{minipage}{0.49\linewidth}
        \centering
        \includegraphics[width=\linewidth]{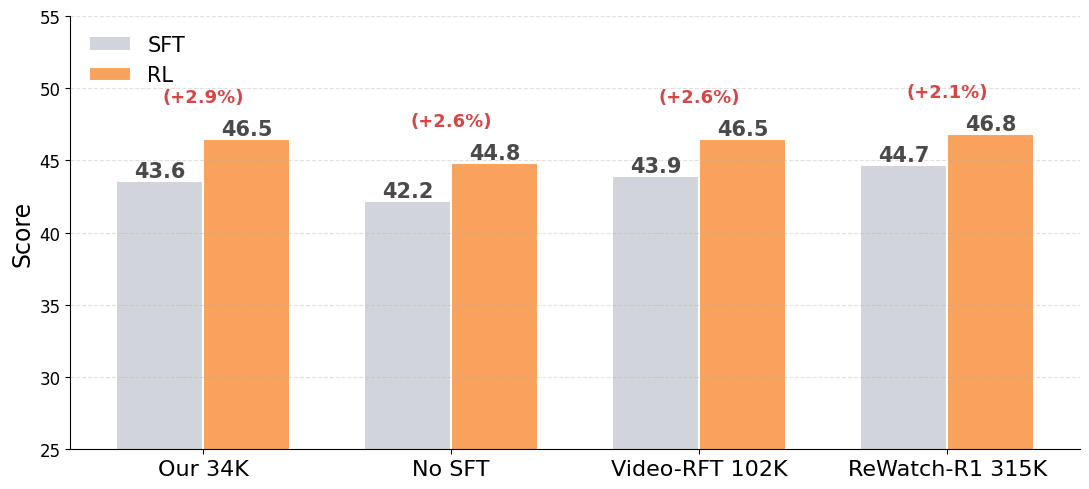} 
        \subcaption{A comparison of Cold Start Stage} 
        \label{fig:abl_cold_satart}
    \end{minipage}
    \begin{minipage}{0.49\linewidth}
        \centering
        \includegraphics[width=\linewidth]{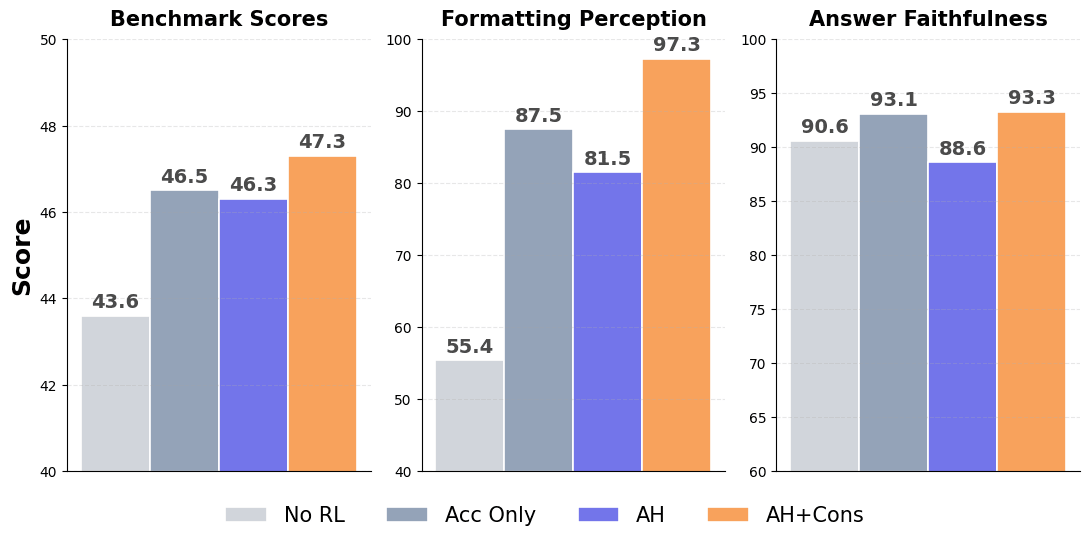} 
        \subcaption{A comparison of the RL stage}
        \label{fig:abl_rl}
    \end{minipage}
    
    \caption{Ablation study on cold-start data and reward signals. Benchmark scores average reasoning and general benchmarks. External Evidence and Answer Faithfulness are evaluated by a closed-source large model, which, given a question and an answer, determines whether all visual information is present within the perceptual unit and whether the answer is consistent with the perceptual results.}
    \label{fig:posttrain_stats}
\end{figure}
\noindent\textbf{Cold-Start Stage:}
As illustrated in Fig.~\ref{fig:abl_cold_satart}, our cold-start data is substantially smaller than that of other methods. Unsurprisingly, the performance after SFT alone is proportional to this limited data scale. However, although our model shows only modest gains after SFT, it achieves comparable performance to other methods after RL training. Our approach demonstrates that RL rather than SFT is the primary driver of capability improvement, while cold-start data mainly teaches the reasoning format.

\noindent\textbf{Reinforcement Learning Stage:}
As shown in Fig.~\ref{fig:abl_rl}, we observe that incorporating the anti-hallucination reward alone does not directly improve model capability. This reveals an important limitation of the hallucination reward, as it encourages conservative evidence generation rather than richer perception.
As illustrated in the \emph{External Evidence} section, both the baseline without RL and the variant with only the anti-hallucination reward underperform compared to using only the accuracy reward. However, adding the consistency reward nearly eliminates out-of-format perception outputs, enabling the anti-hallucination reward to function effectively. Furthermore, the consistency reward substantially improves the alignment between reasoning answers and video evidence, thereby enhancing the model's reasoning capability. These findings demonstrate that the Hallucination reward alone is insufficient, and the structural consistency reward is required to make them effective.

\subsection{Perception Behavior Analysis}

\begin{figure}[t]
  \centering
    \begin{minipage}{0.33\linewidth}
        \centering
        \includegraphics[width=\linewidth]{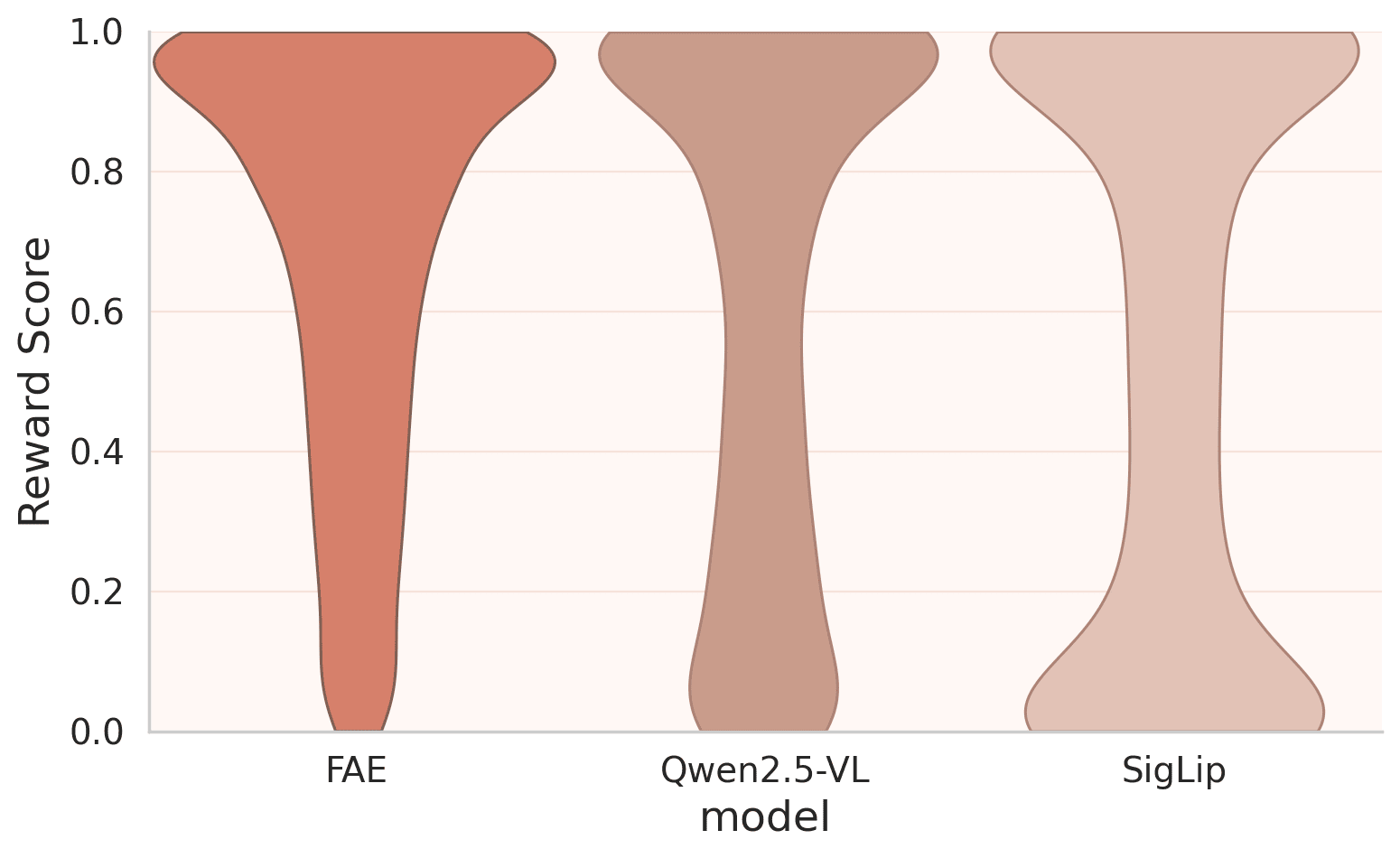} 
        \caption{A comparison of Correct Reward Distribution} 
        \label{fig:reward_score}
    \end{minipage}
     \begin{minipage}{0.3\linewidth}
        \centering
        \includegraphics[width=\linewidth]{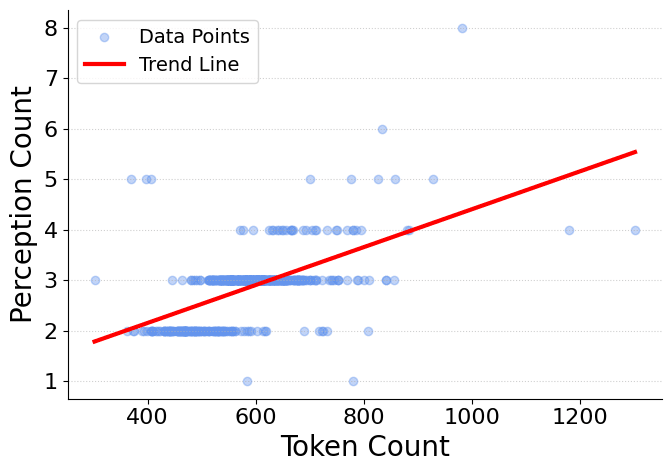}
        \caption{Perception-Token Relationship}
        \label{fig:evid_length}
    \end{minipage}
    \begin{minipage}{0.33\linewidth}
        \centering
        \includegraphics[width=\linewidth]{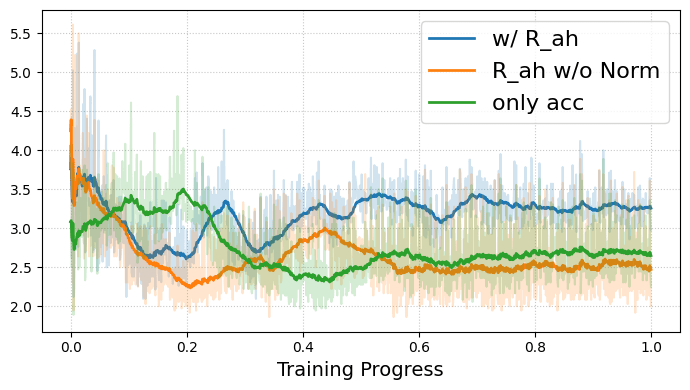}
        \caption{Records of training perception times under different $R_{ah}$ settings}
        \label{fig:evidence_step}
    \end{minipage}
   
\end{figure}

\noindent\textbf{FAE has Higher Differentiation:}
We find that FAE has a larger gap in reward distribution between correct and incorrect samples than the base model, as shown in Figure ~\ref{fig:reward_score}, and produces fewer extreme errors. This suggests that FAE provides a more reliable signal to distinguish between hallucinatory and non-hallucinatory evidence, which is crucial for effective reinforcement learning in video inference tasks. We additionally compared SigLip, which VideoRFT uses for evaluation. SigLIP shows the lowest discriminability among the three. This suggests that generic vision encoders lack sufficient discrimination for hallucination evaluation, highlighting the necessity of specialized factuality evaluators.

\noindent\textbf{Token-Perception Correlation:}
Fig.~\ref{fig:evid_length} reveals a strong positive correlation between generated tokens and valid evidence segments. This demonstrates that Longer reasoning outputs correspond to richer perception rather than redundant text, indicating that the model actively gathers visual evidence. The model uses longer outputs to gather more visual evidence, ensuring reasoning is grounded in observed facts rather than language priors.

\noindent\textbf{Reward Hacking in Perception Rewards:} We find that the anti-hallucination reward can be exploited. Since the reward averages scores across evidence units, the model may selectively provide only high-confidence evidence to maximize reward, resulting in fewer perception steps. Fig.~\ref{fig:evidence_step} shows that without nonlinear normalization, perception steps decrease during training as the model learns to reduce evidence generation. With normalization, the model produces more evidence while preventing exploitation. Combined with our finding that anti-hallucination reward alone induces conservative strategies, both nonlinear normalization and the consistency reward are necessary to mitigate reward hacking.

\section{Conclusion}

In this paper, we study how to improve the reliability of video reasoning by explicitly supervising intermediate perceptual evidence. By introducing perception-aware rewards in Video-DPL, our framework encourages models to produce grounded visual evidence and align reasoning with observed video content. Experiments demonstrate that reliable supervision over perceptual evidence significantly improves reasoning performance across video understanding benchmarks. This finding suggests that strengthening perception reliability is a critical step toward robust video reasoning in VideoLLMs.
For future work, we plan to develop quantifiable hallucination evaluation models and extend perception rewards to tool-use frameworks for video reasoning.

\bibliographystyle{splncs04}
\bibliography{main}

\clearpage
\setcounter{page}{1}

\end{document}